\documentclass[journal]{IEEEtran} 
\IEEEoverridecommandlockouts
\usepackage{amsmath,amssymb,amsfonts}
\usepackage{hyperref}
\hypersetup{
    colorlinks=true,
    linkcolor=black,
    filecolor=black,      
    urlcolor=cyan,
    citecolor=black,
    pdftitle={Overleaf Example},
    pdfpagemode=FullScreen,
    }
\usepackage[utf8]{inputenc}
\usepackage[english]{babel}
\usepackage{algorithmic}
\usepackage{graphicx}
\usepackage{textcomp}
\usepackage{xcolor}
\usepackage[backend=biber,style=ieee,sorting=none]{biblatex}
\usepackage{csquotes}
\usepackage{svg}
\usepackage{subcaption}
\DeclareGraphicsExtensions{.pdf,.jpg,.png}
\IEEEoverridecommandlockouts 

\usepackage{xcolor}

\title{Virtually turning robotic manipulators into worn devices: opening new horizons for wearable assistive robotics}
\author{\IEEEauthorblockN{Alexis Poignant,
Nathanaël Jarrassé and 
Guillaume Morel}\\
\IEEEauthorblockA{CNRS, UMR 7222, ISIR / INSERM, U1150 Agathe‑ISIR, Sorbonne Université, 75005 Paris, France. }}

\begin{document}

\maketitle

\begin{abstract}
Robotic sensorimotor extensions (supernumerary limbs, prosthesis, handheld tools) are worn devices used to interact with the nearby environment, whether to assist the capabilities of impaired users or to enhance the dexterity of industrial operators. Despite numerous mechanical achievements, embedding these robotics devices remains critical due to their weight and discomfort. To emancipate from these mechanical constraints, we propose a new hybrid system using a virtually worn robotic arm in augmented-reality, and a real robotic manipulator servoed on such virtual representation. We aim at bringing an illusion of wearing a robotic system while its weight is fully deported, thinking that this approach could open new horizons for the study of wearable robotics without any intrinsic impairment of the human movement abilities.
\end{abstract}

\IEEEpeerreviewmaketitle

A video showing the set-ups and some of their applications can be found on YouTube at: \url{https://youtu.be/EgwzT784Fws}, and we will be referring to it throughout the paper.

\section{Introduction}

Among wearable robotic devices, robotic sensorimotor extensions are serial augmentations of the human body aiming to increase the user's motor abilities. These devices include but does not limit to prosthesis \cite{ribeiro_analysis_2019}, supernumerary limbs \cite{yang_supernumerary_2021} or handheld robotic tools \cite{gregg-smith_design_2015}. Despite mechanical advances over the last decades, embedding these robotic devices on the human body remains critical as their weight \cite{handy2018satisfaction, prattichizzo_human_2021}, cumbersomeness and limited functionalities \cite{cordella_literature_2016} restrain their comfort and acceptance \cite{martinez2021wearable}, leading many potential users to prefer simple mechanical devices \cite{millstein_prosthetic_1986} or the use of their natural limbs \cite{guggenheim_laying_2020}.

In spite of these disadvantages, those wearable devices remain unique: as indicated by their name, they are worn by their user, meaning that, in addition to being transportable, the robot end-effector's transform relative to the world frame $T_{E_R \to W}$ can be written as a composition of: 

\begin{itemize}
    \item the robotic control relative to its human attachment base $T_{E_R \to W}$; 
    \item and the movement of the human body relative to the world $T_{E_H \to W}$.
\end{itemize}
We can rephrase it as: the robot and the human kinematic chains are serially connected. It writes:

\begin{equation}
    T_{E_R \to W} = T_{E_R \to E_H}  T_{E_H \to W} ~,
\end{equation}

where $T_{A\to B}$ the transform from frame $A$ to frame $B$, $E_R$ the robotic end-effector frame, $E_H$ the human effector frame to which the robot base is attached, and $W$ the world frame.

\begin{figure}[h!]
    \centering
    \includegraphics[width=0.48\textwidth]{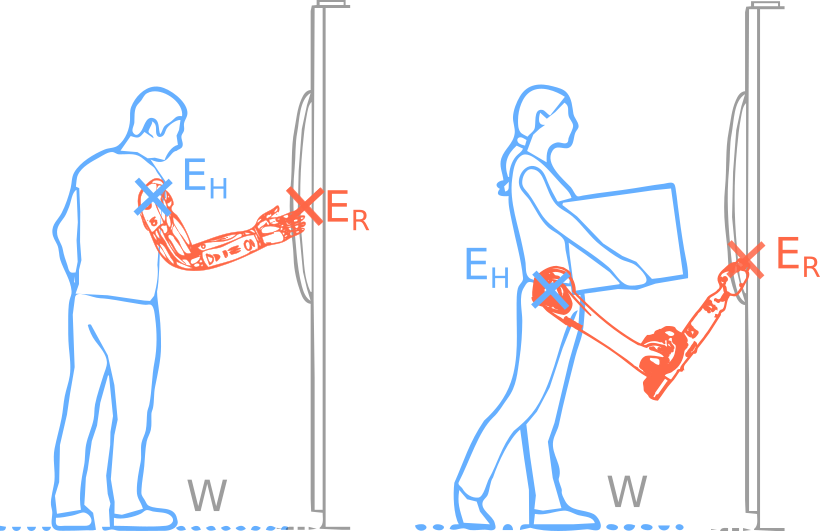}
    \vspace{2mm}
    \caption{A robotic sensorimotor extension in orange (prosthesis or supernumerary robotic limb) with effector $E_R$ (orange cross) is attached on an user at $E_H$ (blue cross), to perform a reaching task.}
    \label{fig:compensations}
\end{figure}

If the human plus the robot system form a kinematically redundant chain(i.e. when the dimension of joint space is greater than the dimension of end-effector space \cite{conkur_clarifying_1997}), it also means that the task completion is shared between the human motion $T_{E_H \to W}$ and the robotic motion $T_{E_R \to E_H}$ (Fig.~\ref{fig:compensations}):
\begin{itemize}
    \item $T_{E_R \to E_H}$ is usually governed by an auxiliary signal coming from a joystick-like input \cite{zahraee_toward_2010}, a sensor system measuring electrophysiological activities (such as muscles activities through EMG \cite{li_quantifying_2010} or cerebral activities through EEG \cite{vilela_applications_2020}), or an automatic task-oriented behaviour \cite{luo_modeling_2021};
    \item $T_{E_H \to W}$ depends on the attachment point of the worn device, whether the arm or forearm for an upper-limb prosthesis, the trunk or hips for a supernumerary limb, or the hand for an handheld tool, and of the movements of the user. 
\end{itemize}

Yet, this kinematic redundancy is frequently underused, leading to undesirable postures for the user: 
\begin{itemize}
    \item in prosthesis control, $T_{E_R \to E_H}$ is piloted by myoelectric activities of the stump and is often slow and non-intuitive leading to important body compensations \cite{hussaini_categorization_2017} by $T_{E_H \to W}$ over $T_{E_R \to E_H}$;
    
    \item oppositely, in supernumerary robotic limbs, the weight of the robotic device limits the dexterity \cite{shimobayashi_independent_2021} of $T_{E_H \to W}$ (by creating imbalance and fatigue) and imposes the use of $T_{E_R \to E_H}$ over $T_{E_H \to W}$.
    
\end{itemize}

Currently, the discomfort and limited performances of these devices and of their control still limit the potential of their usage, as the weight of the device is fully carried by the user.

% \begin{figure}[t!]
%     \label{fig:setups}
%     \centering
%     \includegraphics[width=\linewidth]{worn_vs_set_kine.png}
%     \caption{Kinematic representation of a robotic manipulator set on table and a human(left), and a worn robot (right). The set case is constituted of two parallel chains, the human (blue) constituted of various effectors (hands, mouth, feet), and the robot (orange). In the worn scenario, the human and robot are serially connected. Nécessaire ?}
%     \label{fig:worn_vs_set}
% \end{figure}

\section{Virtually turning robotic manipulators into worn devices}
\begin{figure}[t!]
    \centering
    \includegraphics[width=0.9\linewidth]{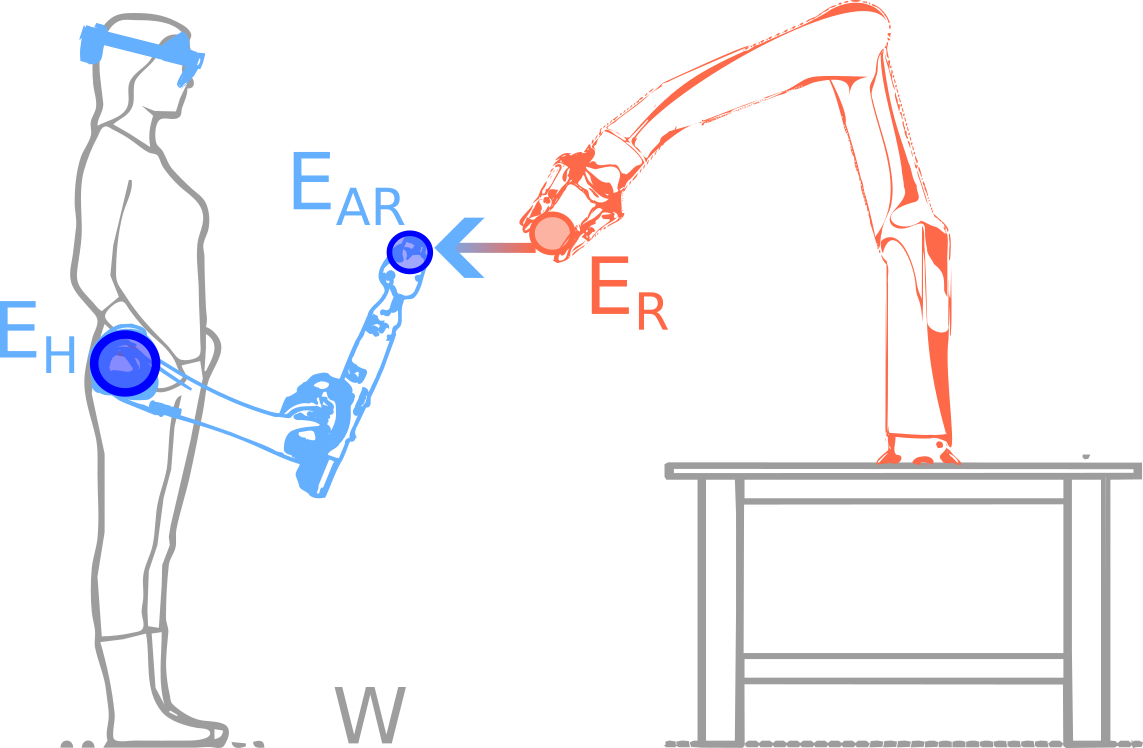}
    \vspace{5mm}
    \caption{The user (grey) wears a virtual robotic arm in blue at attach point $E_H$ (blue circle on the user), with virtual end-effector $E_{AR}$ (blue circle at the end of the robot). The virtual arm made visible thanks to the AR headset in blue on the users' head. Then, the virtual end-effector servoes the effector of a deported robotic arm's (orange circle) set on a table, with a delay depending on the robot dynamics and control (symbolized by the arrow).}
    \label{fig:setups}
\end{figure}

To solve the previous issue, we propose a new set-up which fully deports the weight of the robotic device to a nearby environment, while still keeping a serial connection between the user and the device: we create an augmented-reality representation of an arm, attached to the body frame $E_H$, ending with a virtual end-effector frame $E_{AR}$, while a robotic manipulator set nearby with an end-effector frame $E_R$ is servoed to follow $E_{AR}$ (see Fig.~\ref{fig:setups} and video Sec.~1):

\begin{equation}
    \begin{cases}
      T_{E_{AR} \to W} = T_{E_{AR} \to E_H}  T_{E_H \to W}\\
      T_{E_R \to W}(t) = f(T_{E_{AR} \to W}(t))
    \end{cases}
\end{equation}
with $f$ the robot dynamics and control, causing delay when visually servoing the position of robotic end-effector. 

The fundamental serial connection of worn robotic devices is now purely virtual, allowing to deport the weight, whereas the real robotic manipulator, physically detached from the human body, performs the desired task with a certain delay (imposed by its dynamics or control loop limitations). By doing so, we aim at virtually creating a sensation of physical embodiment of the deported manipulator.

As said, the robotic manipulator is often slower than the human motions $T_{E_H \to W}$, creating a delay between the virtual frame $E_{AR}$ motions and the robotic effector $E_R$ motions. But, by representing the virtual arm in augmented-reality, we partially solve these dynamic issues as we delete the perception delay for the user, tending to more stable motions. This representation is however not necessary for the system to actually work and, with training, can be removed (see video Sec.~4). 

Finally, $T_{E_H \to W}$ can be generated from any body part $E_H$, such as the head, trunk or arm (see video Sec.~2) depending on the user's preference and the applications, while $T_{E_{AR} \to E_H}$ can still be controlled by an auxiliary input, like a joystick (see video Sec.~3).

To sum up, the proposed set-up allows to:
\begin{itemize}
    \item shift the weight and embody any existing robotic system as a virtually worn device, even heavyweight devices, offering the possibility to reach higher robotic performances and higher comfort;
    \item create mechanically unfeasible limbs, such as an arm attach to the head for disability assistive applications, or a extended third arm with wide workspace for industrial applications;
    \item instantly attach and detach any robot device to the user's body, switch systems and shift the attach point from one to another for different use-case scenarios, unlike current devices which usually require important preparation times.
\end{itemize}

\section{Practical implementation}
\subsection{Controlled Environment}
\begin{figure}[t!]
    \centering
    \includegraphics[width=0.75\linewidth]{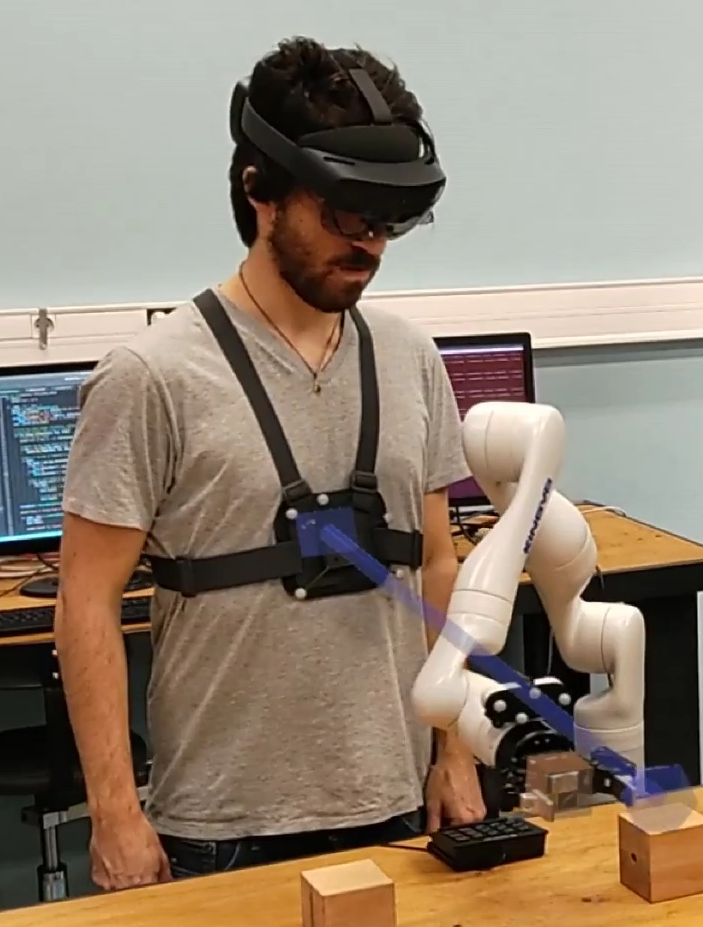}
    \vspace{5mm}
    \caption{The user frame $E_H$ (blue cube) and robot end-effector frame $E_R$ are tracked using an Optitrack system. The Microsoft Hololens V2 headset displays the virtual arm (blue), its virtual effector $E_{AR}$ (blue octagon) and current robot arm effector configuration (orange cube), represented here. The robot effector is servoed to follow the virtual effector.}
    \label{fig:optitrack set_up}
\end{figure}

For research purposes, in controlled and room-sized environments, the easiest and most flexible implementation is to use an optical motion capture system such as the Optitrack system, and an augmented-reality headset like the Microsoft Hololens V2 for display, as seen Fig.~\ref{fig:optitrack set_up}.

The motion capture system allows to capture the body transform $T_{E_H \to W}$ while easily switching from one body attach point $E_H$ to another. By putting another marker on the robot end-effector $E_R$ it also allows to perform visual position servoing. The augmented-reality headset is then only used for display and is actually not necessary to control the robot, rather serving as a visual feedback provider which increases the easiness of use. Finally, another auxiliary control input, such as a joystick, button or electrophysiological system can be added, whether to reconfigure and control the virtual linkage $T_{E_{AR} \to E_H}$ or to control a robotic gripper.

\subsection{Fully wearable set-up for applied scenarios}
Optical motion capture systems cannot be worn. However, the Microsoft Hololens V2 is equipped with a visual 3D slam function which allows to detect at all time the position of the user's head and can be used to calibrate the robot with the headset. By using Inertial Measurement Units on the trunk and the arm, and a simple kinematic model of the user, one can then reconstruct the upper-body joint positions and the transform $T_{E_H \to W}$ (see video Sec.~5). Auxiliary inputs measurement can also be implemented in the headset, using its gaze tracker to control the gripper for example.

This measurement system is fully wearable and transportable (the headset being autonomous) and cheaper than the optical motion capture system, but it requires more development and time to be fully operational, as well as previous knowledge regarding the user's body part to which the virtual robotic arm should be attached.

However, as it can be seen on our setup, a critical aspect of worn robotics which was not resolved here is the transportability of the system, as the robotic manipulator is set on a table. To solve the latter, we could consider to embed the manipulator on a mobile base, wheeled or legged, which could follow the user's motions in a larger environment, or on a wheelchair for assistive applications, and use an inverse kinematic model of both the base and the manipulator to servo the system.

\section{Conclusion}
We proposed a novel approach for robotic extensions, in which the worn device is fully virtual but distantly controls a real robotic manipulator. This allows to virtually embody any high-performance robotic device without weight or embodiment issues, and explore applications that where previously unfeasible — an arm attached to the head or a very long arm fixed to the trunk — without constraining the user's body with heavy and discomfortable wearable devices. We think that this approach opens new horizons for the study or wearable robotics by fully emancipating from the mechanical constraints of the systems, allowing to explore both fundamental and applied aspects of worn devices without intrinsic impairment of the human motions.

\hypersetup{urlcolor = black}
\printbibliography

@article{millstein_prosthetic_1986,
  title={Prosthetic use in adult upper limb amputees: a comparison of the body powered and electrically powered prostheses},
  author={Millstein, SG and Heger, H and Hunter, GA},
  journal={Prosthetics and orthotics international},
  volume={10},
  number={1},
  pages={27--34},
  year={1986},
  publisher={SAGE Publications Sage UK: London, England}
}

@article{guggenheim_laying_2020,
	title = {Laying the Groundwork for Intra-Robotic-Natural Limb Coordination: Is Fully Manual Control Viable?},
	volume = {9},
	issn = {2573-9522},
	url = {https://dl.acm.org/doi/10.1145/3377329},
	doi = {10.1145/3377329},
	shorttitle = {Laying the Groundwork for Intra-Robotic-Natural Limb Coordination},
	pages = {1--12},
	number = {3},
	journaltitle = {{ACM} Transactions on Human-Robot Interaction},
	shortjournal = {J. Hum.-Robot Interact.},
	author = {Guggenheim, Jacob W. and Parietti, Federico and Flash, Tamar and Asada, H. Harry},
	urldate = {2022-10-21},
	date = {2020-09-30},
	langid = {english},
}

@article{ribeiro_analysis_2019,
	title = {Analysis of Man-Machine Interfaces in Upper-Limb Prosthesis: A Review},
	volume = {8},
	issn = {2218-6581},
	url = {http://www.mdpi.com/2218-6581/8/1/16},
	doi = {10.3390/robotics8010016},
	shorttitle = {Analysis of Man-Machine Interfaces in Upper-Limb Prosthesis},
	pages = {16},
	number = {1},
	journaltitle = {Robotics},
	shortjournal = {Robotics},
	author = {Ribeiro, José and Mota, Francisco and Cavalcante, Tarique and Nogueira, Ingrid and Gondim, Victor and Albuquerque, Victor and Alexandria, Auzuir},
	urldate = {2022-02-28},
	date = {2019-02-21},
	langid = {english},
}

@article{prattichizzo_human_2021,
	title = {Human augmentation by wearable supernumerary robotic limbs: review and perspectives},
	volume = {3},
	issn = {2516-1091},
	url = {https://iopscience.iop.org/article/10.1088/2516-1091/ac2294},
	doi = {10.1088/2516-1091/ac2294},
	shorttitle = {Human augmentation by wearable supernumerary robotic limbs},
	pages = {042005},
	number = {4},
	journaltitle = {Progress in Biomedical Engineering},
	shortjournal = {Prog. Biomed. Eng.},
	author = {Prattichizzo, Domenico and Pozzi, Maria and Lisini Baldi, Tommaso and Malvezzi, Monica and Hussain, Irfan and Rossi, Simone and Salvietti, Gionata},
	urldate = {2022-02-28},
	date = {2021-10-01},
	langid = {english},
}

@article{yang_supernumerary_2021,
	title = {Supernumerary Robotic Limbs: A Review and Future Outlook},
	volume = {3},
	issn = {2576-3202},
	url = {https://ieeexplore.ieee.org/document/9446590/},
	doi = {10.1109/TMRB.2021.3086016},
	shorttitle = {Supernumerary Robotic Limbs},
	pages = {623--639},
	number = {3},
	journaltitle = {{IEEE} Transactions on Medical Robotics and Bionics},
	shortjournal = {{IEEE} Trans. Med. Robot. Bionics},
	author = {Yang, Bo and Huang, Jian and Chen, Xinxing and Xiong, Caihua and Hasegawa, Yasuhisa},
	urldate = {2022-02-28},
	date = {2021-08},
	langid = {english},
}

@inproceedings{gregg-smith_design_2015,
	location = {Seattle, {WA}, {USA}},
	title = {The design and evaluation of a cooperative handheld robot},
	isbn = {978-1-4799-6923-4},
	url = {http://ieeexplore.ieee.org/document/7139456/},
	doi = {10.1109/ICRA.2015.7139456},
	eventtitle = {2015 {IEEE} International Conference on Robotics and Automation ({ICRA})},
	pages = {1968--1975},
	booktitle = {2015 {IEEE} International Conference on Robotics and Automation ({ICRA})},
	publisher = {{IEEE}},
	author = {Gregg-Smith, Austin and Mayol-Cuevas, Walterio W.},
	urldate = {2022-03-04},
	date = {2015-05},
	langid = {english},
}

@article{zahraee_toward_2010,
	title = {Toward the Development of a Hand-Held Surgical Robot for Laparoscopy},
	issn = {1083-4435, 1941-014X},
	url = {http://ieeexplore.ieee.org/document/5523951/},
	doi = {10.1109/TMECH.2010.2055577},
	pages = {5523951},
	journaltitle = {{IEEE}/{ASME} Transactions on Mechatronics},
	shortjournal = {{IEEE}/{ASME} Trans. Mechatron.},
	author = {Zahraee, Ali Hassan and Paik, Jamie Kyujin and Szewczyk, Jerome and Morel, Guillaume},
	urldate = {2022-03-04},
	date = {2010-12},
	langid = {english},
}

@article{martinez2021wearable,
  title={Wearable Assistive Robotics: A Perspective on Current Challenges and Future Trends},
  author={Martinez-Hernandez, Uriel and Metcalfe, Benjamin and Assaf, Tareq and Jabban, Leen and Male, James and Zhang, Dingguo},
  journal={Sensors},
  volume={21},
  number={20},
  pages={6751},
  year={2021},
  publisher={MDPI}
}

@article{handy2018satisfaction,
  title={Satisfaction of patients with amputated lower limb wearing external prostheses},
  author={Handy Eone, D and Nseme Etouckey, E and Essi, MJ and Ngo Nyemb, TM and Ngo Nonga, B and Ibrahima, F and Sosso, MA},
  journal={International Journal of Orthopaedics},
  volume={4},
  number={1},
  pages={368--372},
  year={2018}
}

@inproceedings{shimobayashi_independent_2021,
	location = {Rovaniemi Finland},
	title = {Independent Control of Supernumerary Appendages Exploiting Upper Limb Redundancy},
	isbn = {978-1-4503-8428-5},
	url = {https://dl.acm.org/doi/10.1145/3458709.3458980},
	doi = {10.1145/3458709.3458980},
	eventtitle = {{AHs} '21: Augmented Humans International Conference 2021},
	pages = {19--30},
	booktitle = {Augmented Humans Conference 2021},
	publisher = {{ACM}},
	author = {Shimobayashi, Hideki and Sasaki, Tomoya and Horie, Arata and Arakawa, Riku and Kashino, Zendai and Inami, Masahiko},
	urldate = {2022-03-04},
	date = {2021-02-22},
	langid = {english},
}

@article{cordella_literature_2016,
	title = {Literature Review on Needs of Upper Limb Prosthesis Users},
	volume = {10},
	issn = {1662-453X},
	url = {http://journal.frontiersin.org/Article/10.3389/fnins.2016.00209/abstract},
	doi = {10.3389/fnins.2016.00209},
	journaltitle = {Frontiers in Neuroscience},
	shortjournal = {Front. Neurosci.},
	author = {Cordella, Francesca and Ciancio, Anna Lisa and Sacchetti, Rinaldo and Davalli, Angelo and Cutti, Andrea Giovanni and Guglielmelli, Eugenio and Zollo, Loredana},
	urldate = {2022-10-18},
	date = {2016-05-12},
	langid = {english},
}

@article{hussaini_categorization_2017,
	title = {Categorization of compensatory motions in transradial myoelectric prosthesis users},
	volume = {41},
	issn = {0309-3646},
	url = {https://journals.lww.com/00006479-201741030-00009},
	doi = {10.1177/0309364616660248},
	pages = {286--293},
	number = {3},
	journaltitle = {Prosthetics \& Orthotics International},
	author = {Hussaini, Ali and Zinck, Arthur and Kyberd, Peter},
	urldate = {2022-03-04},
	date = {2017-06},
	langid = {english},
}

@article{li_quantifying_2010,
	title = {Quantifying Pattern Recognition—Based Myoelectric Control of Multifunctional Transradial Prostheses},
	volume = {18},
	issn = {1534-4320, 1558-0210},
	url = {https://ieeexplore.ieee.org/document/5378627/},
	doi = {10.1109/TNSRE.2009.2039619},
	pages = {185--192},
	number = {2},
	journaltitle = {{IEEE} Transactions on Neural Systems and Rehabilitation Engineering},
	shortjournal = {{IEEE} Trans. Neural Syst. Rehabil. Eng.},
	author = {Li, Guanglin and Schultz, Aimee E. and Kuiken, Todd A.},
	urldate = {2022-03-04},
	date = {2010-04},
	langid = {english},
}

@incollection{vilela_applications_2020,
	title = {Applications of brain-computer interfaces to the control of robotic and prosthetic arms},
	volume = {168},
	isbn = {978-0-444-63934-9},
	url = {https://linkinghub.elsevier.com/retrieve/pii/B9780444639349000081},
	pages = {87--99},
	booktitle = {Handbook of Clinical Neurology},
	publisher = {Elsevier},
	author = {Vilela, Marco and Hochberg, Leigh R.},
	urldate = {2022-03-04},
	date = {2020},
	langid = {english},
	doi = {10.1016/B978-0-444-63934-9.00008-1},
}

@article{conkur_clarifying_1997,
	title = {Clarifying the definition of redundancy as used in robotics},
	volume = {15},
	issn = {0263-5747, 1469-8668},
	url = {https://www.cambridge.org/core/product/identifier/S0263574797000672/type/journal_article},
	doi = {10.1017/S0263574797000672},
	pages = {583--586},
	number = {5},
	journaltitle = {Robotica},
	shortjournal = {Robotica},
	author = {Conkur, E. Sahin and Buckingham, Rob},
	urldate = {2022-10-28},
	date = {1997-09},
	langid = {english},
}

@article{luo_modeling_2021,
	title = {Modeling and Balance Control of Supernumerary Robotic Limb for Overhead Tasks},
	volume = {6},
	issn = {2377-3766, 2377-3774},
	url = {https://ieeexplore.ieee.org/document/9384151/},
	doi = {10.1109/LRA.2021.3067850},
	pages = {4125--4132},
	number = {2},
	journaltitle = {{IEEE} Robotics and Automation Letters},
	shortjournal = {{IEEE} Robot. Autom. Lett.},
	author = {Luo, Jianwen and Gong, Zelin and Su, Yao and Ruan, Lecheng and Zhao, Ye and Asada, H. Harry and Fu, Chenglong},
	urldate = {2022-10-26},
	date = {2021-04},
	langid = {english},
}

\end{document}